\documentclass[lettersize,journal]{IEEEtran}
\date{April 2026}
\usepackage{amsmath,amsfonts}
\usepackage{algorithmic}
\usepackage{algorithm}
\usepackage{amssymb}
\usepackage{array}
\usepackage[caption=false,font=normalsize,labelfont=sf,textfont=sf]{subfig}
\usepackage{textcomp}
\usepackage{stfloats}
\usepackage{url}
\usepackage{verbatim}
\usepackage{multirow}
\usepackage{graphicx}
\usepackage{cite}
\begin{document}

\title{HTT-Net: Hierarchical Text-guided Transition Modeling for Surgical Video Phase Recognition}

\author{Kunjie~Deng,
        Jinghui~Zhang,
        Weidong~Chen,~\IEEEmembership{Member,~IEEE},
        Ganbin~Li,
        Xiangjun~Lyu,
        Zhendong~Mao,~\IEEEmembership{Member,~IEEE},
        and Yingchi~Yang%
\thanks{Manuscript received XX XX, 2026; revised XX XX, 2026.}
\thanks{Kunjie Deng and Jinghui Zhang contributed equally to this work. Corresponding authors: Weidong Chen, Xiangjun Lyu, and Yingchi Yang.}
\thanks{Kunjie Deng, Weidong Chen, and Zhendong Mao are with the School of Information Science and Technology, University of Science and Technology of China, Hefei, China (e-mail: iideng@mail.ustc.edu.cn; chenweidong@ustc.edu.cn; zdmao@ustc.edu.cn).}
\thanks{Jinghui Zhang, Ganbin Li, and Yingchi Yang are with the Department of General Surgery, Beijing Friendship Hospital, Capital Medical University, Beijing 100050, China; State Key Lab of Digestive Health, Beijing 100050, China; and National Clinical Research Center for Digestive Diseases, Beijing 100050, China (e-mail: jinghuizhang@mail.ccmu.edu.cn; ligb0808@163.com; yangyingchi@ccmu.edu.cn).}
\thanks{Xiangjun Lyu is with the Senior Department of Urology, Chinese PLA General Hospital, Beijing 100039, China (e-mail: lyuxiangjun@163.com).}}

\markboth{IEEE Transactions on Medical Imaging,~Vol.~XX, No.~XX, 2026}%
{Author \MakeLowercase{\textit{et al.}}: HTT-Net}

\maketitle

\begin{abstract}
Surgical video phase recognition is a fundamental task in computer-assisted intervention, supporting workflow understanding, intraoperative guidance, and surgical quality assessment. Although recent visual-temporal models have achieved promising progress, accurate and temporally coherent phase recognition remains challenging due to local visual ambiguity, transient prediction noise, and insufficient use of procedural semantics. To address these challenges, we propose \textbf{HTT-Net}, a \textbf{H}ierarchical \textbf{T}ext-guided \textbf{T}ransition modeling \textbf{N}etwork for surgical video phase recognition. The key idea is to introduce structured surgical semantic knowledge into phase-aware segment construction and semantic refinement. Specifically, we construct a \textbf{hierarchical surgical semantic memory} with intra-phase descriptions, inter-phase transition descriptions, and fine-grained semantic units. Based on this memory, the proposed \textbf{Transition-Aware Segment Construction} (TAS-Con) organizes frame-level evidence into coherent segment representations and handles boundary clips with inter-phase transition descriptions. Furthermore, we introduce \textbf{Transition-Aware Segment Calibration} (TAS-Calib), which calibrates phase-aware segment representations through hierarchical surgical semantics and improves discrimination under visual ambiguity without dense frame-level vision-language fusion. Experiments on Cholec80 and LCRS-100 demonstrate the effectiveness of HTT-Net for robust surgical video phase recognition.
\end{abstract}

\begin{IEEEkeywords}
Hierarchical semantic guidance, transition-aware segment construction, surgical video phase recognition, surgical workflow analysis, vision-language learning.
\end{IEEEkeywords}

\section{Introduction}

\begin{figure}[!t]
    \centering
    \includegraphics[
        width=\linewidth,
        trim=2.5in 0.8in 2.5in 0.5in,
        clip
    ]{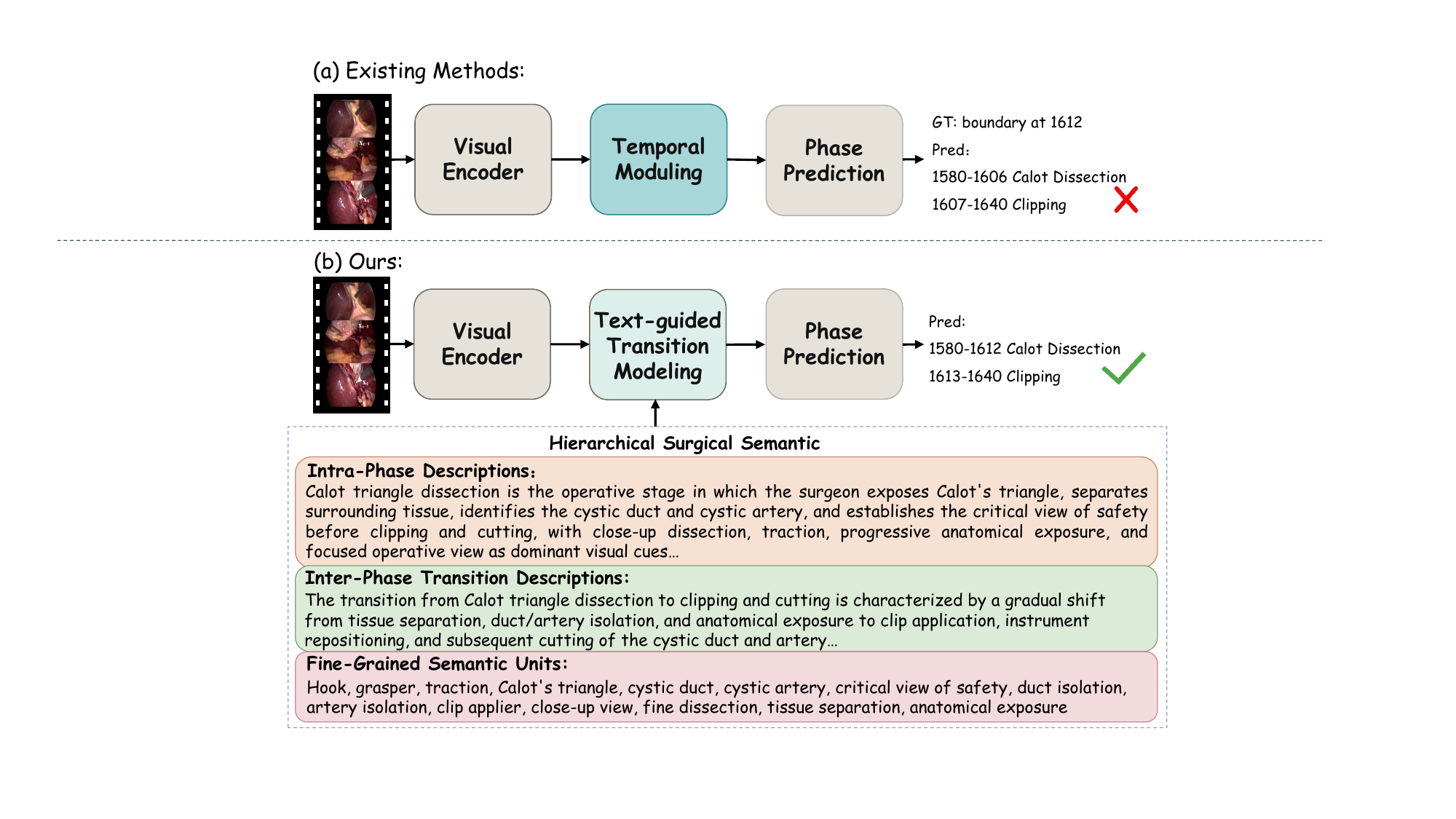}
    \caption{Conceptual motivation of text-guided transition-aware surgical video phase recognition.
    (a) Conventional visual-temporal modeling relies on visual features and temporal modules for phase prediction, which may be affected by ambiguous boundary evidence.
    (b) The proposed text-guided transition modeling introduces hierarchical surgical semantic memory after visual feature extraction, including intra-phase descriptions, inter-phase transition descriptions, and fine-grained semantic units, to provide semantic guidance for more reliable phase prediction.}
    \label{fig:intro_motivation}
\end{figure}

\IEEEPARstart{S}{urgical} video phase recognition assigns phase labels to surgical video frames and describes the temporal progression of an operation.
As a fundamental task in surgical workflow analysis, it supports computer-assisted intervention, intraoperative guidance, surgical skill assessment, operating room management, and postoperative quality analysis~\cite{padoy2012statistical,neumuth2017process,garrow2021systematic,demir2023review}.
Accurate and temporally coherent recognition is therefore important for reliable intelligent surgical systems.

Early methods relied on handcrafted visual features and statistical temporal models, such as hidden Markov models, dynamic time warping, and knowledge-driven workflow modeling~\cite{ahmadi2006recovery,blum2008workflow,padoy2012statistical,katic2014knowledge}.
Recent deep learning methods typically follow a visual-temporal modeling paradigm.
As illustrated in Fig.~\ref{fig:intro_motivation}(a), frames are first processed by a spatial module to extract visual features, and a temporal module then models workflow dynamics for dense phase prediction~\cite{twinanda2016endonet,jin2017svrcnet,czempiel2020tecno,jin2021tmrnet,gao2021transsvnet,yue2023cmtnet,liu2023skit,yang2024surgformer}.
Recent studies further improve this paradigm through long-video modeling, state-space modeling, parameter-efficient transfer, clip-aware temporal reasoning, and query-based interaction~\cite{liu2025lovit,cao2024srmamba,yang2026surgpetl,yang2025dacat,zhang2026b2qnet}.
However, these methods still rely mainly on visual evidence and weakly explore procedural semantics.

A key challenge is that surgical clips have different phase compositions.
Most clips are dominated by a single phase and should form stable phase-consistent segments, whereas boundary clips may contain adjacent phases with mixed visual cues.
Since surgical transitions are often gradual, preliminary frame-wise predictions may suffer from premature switching, delayed transitions, short-term oscillations, or fragmented segments.
Local appearance changes caused by tool motion, smoke, occlusion, bleeding, or camera movement may also be misinterpreted as phase changes.
Thus, the key problem is to identify meaningful procedural transitions and organize frames into semantically coherent phase-aware segments.

Another limitation is the insufficient use of explicit surgical semantic knowledge.
Surgical phases are associated with procedural goals, anatomical contexts, instruments, actions, and visual cues, while transitions reflect changes in operative goals, tool usage, anatomical exposure, and local actions.
Such descriptions can provide complementary guidance when visual evidence is ambiguous, especially around phase boundaries or between visually similar phases.
Inspired by vision-language learning~\cite{radford2021clip,gao2023clipadapter,li2023blip2} and surgical video representation studies~\cite{hirsch2023selfsupervised,batic2024endovit,ramesh2023dissecting,nwoye2022rendezvous,yang2026surgpetl}, we introduce structured procedural semantics after visual feature extraction.
As shown in Fig.~\ref{fig:intro_motivation}(b), the proposed framework obtains frame-level visual features through a spatial module and then performs text-guided transition modeling with hierarchical surgical semantic memory.

Motivated by these observations, we propose \textbf{HTT-Net}, a \textbf{H}ierarchical \textbf{T}ext-guided \textbf{T}ransition modeling \textbf{N}etwork for surgical video phase recognition.
We construct a hierarchical surgical semantic memory with intra-phase descriptions, inter-phase transition descriptions, and fine-grained semantic units as phase identity, transition plausibility, and local discrimination references.
Based on this memory, \textbf{Transition-Aware Segment Construction} (TAS-Con) constructs coherent phase-aware segment representations by preserving phase-homogeneous clips and verifying plausible boundary transitions.
Then, \textbf{Transition-Aware Segment Calibration} (TAS-Calib) calibrates compact segment representations through hierarchical surgical semantics, and the calibrated segment context is mapped back to frames for dense phase prediction. The main contributions of this work are summarized as follows:
\begin{itemize}
    \item In this paper, we propose HTT-Net, a hierarchical semantic transition modeling network for surgical video phase recognition. HTT-Net uses hierarchical surgical semantic memory, \emph{i.e.,} intra-phase descriptions, inter-phase transition descriptions, and fine-grained semantic units, as structured procedural knowledge for segment construction and calibration.

    \item We design Transition-Aware Segment Construction (TAS-Con). It preserves phase-consistent clips and suppresses unreliable boundary changes. This produces coherent phase-aware segment representations.

    \item We introduce Transition-Aware Segment Calibration (TAS-Calib). It uses intra-phase, inter-phase transition, and fine-grained text semantics to refine segment representations. This improves discrimination between visually similar phases.

     \item Experiments on Cholec80 and LCRS-100 demonstrate the effectiveness of HTT-Net. Ablation studies further verify the contributions of hierarchical surgical semantic memory, TAS-Con, and TAS-Calib.
\end{itemize}

\section{Related Work}

\subsection{Surgical Video Phase Recognition}

Surgical video phase recognition is a fundamental task in surgical workflow analysis.
Early studies modeled surgical procedures using handcrafted visual cues with statistical or rule-based temporal models, such as hidden Markov models, dynamic time warping, and knowledge-driven workflow modeling~\cite{ahmadi2006recovery,blum2008workflow,padoy2012statistical,katic2014knowledge}.
With deep learning, CNN-based encoders combined with recurrent networks, temporal convolution, memory mechanisms, and Transformer architectures became dominant for visual-temporal representation learning~\cite{twinanda2016endonet,jin2017svrcnet,czempiel2020tecno,jin2021tmrnet,gao2021transsvnet,yue2023cmtnet,liu2023skit,yang2024surgformer,scholar_multiattention_compressed_video,scholar_microaction_imbalance}.
Recent studies further explore hard-frame mining, multi-task learning, vision Transformers, long-range temporal context, end-to-end video learning, state-space modeling, parameter-efficient transfer, clip-aware temporal reasoning, and query-based phase representations~\cite{yi2019hard,jin2020multitask,czempiel2021opera,kiyasseh2023vit,liu2025lovit,rivoir2024bn,cao2024srmamba,yang2026surgpetl,yang2025dacat,zhang2026b2qnet,scholar_graph_moe_time_series,scholar_partially_relevant_video_retrieval,scholar_difference_aware_relation}.
These developments are also related to broader advances in video Transformer modeling~\cite{arnab2021vivit,bertasius2021spacetime}.
However, most existing methods still formulate surgical video phase recognition as visual-temporal modeling, where phase semantics and procedural transition knowledge are only implicitly learned from frame-level annotations.

\subsection{Transition-Aware Semantic Modeling}

Temporal ambiguity around phase boundaries is a common challenge in surgical video phase recognition.
It is related to boundary uncertainty, evaluation sensitivity, and fragmented predictions in dense video understanding~\cite{czempiel2022workflow,funke2023metrics,farha2019mstcn,bahrami2023longterm,lu2024fact,liu2023diffact}.
Existing surgical video phase recognition methods usually address this issue implicitly through recurrent modeling, temporal convolution, Transformer attention, or prediction smoothing~\cite{jin2017svrcnet,czempiel2020tecno,jin2021tmrnet,gao2021transsvnet,yang2024surgformer}.
In contrast, HTT-Net explicitly constructs phase-aware segments and uses transition semantics to distinguish meaningful procedural transitions from transient visual variations.

Vision-language learning shows that textual information can provide semantic supervision beyond discrete labels~\cite{radford2021clip,gao2023clipadapter,li2023blip2,scholar_cascade_crossmodal_segmentation,scholar_weak_text_actor_action,scholar_structured_concepts_captioning,scholar_visual_relationships_captioning}.
In surgical video analysis, semantic knowledge and pretrained representations are valuable because surgical procedures are highly structured~\cite{hirsch2023selfsupervised,batic2024endovit,ramesh2023dissecting,nwoye2022rendezvous,yang2026surgpetl,scholar_video_summary_caption_coherence,scholar_dualpath_emotional_captioning,scholar_emotion_cause_video_captioning,scholar_subjective_objective_captioning,scholar_finegrained_emotion_captioning,scholar_av_exchange_pruning}.
However, most surgical video phase recognition methods treat phase labels as independent categories, while simple class-name prompts or flat textual descriptions cannot capture hierarchical surgical workflow semantics.
This motivates our hierarchical surgical semantic memory, which organizes intra-phase descriptions, inter-phase transition descriptions, and fine-grained semantic units for Transition-Aware Segment Construction and Calibration.
For completeness, additional publications from the same research group on multimodal generation, structured semantics, visual reasoning, and related learning tasks are also cited in this arXiv version~\cite{scholar_bootstrapping_llm_radiology,scholar_creatidesign,scholar_fewshot_multihop_question,scholar_radiology_multigrained,scholar_text_style_transfer,scholar_aspect_sentiment_ccg,scholar_radiology_d2net,scholar_sentiment_live_commenting,scholar_pseudo_word_composed_retrieval,scholar_creatiposter,scholar_stimuli_emotion_adaptor,scholar_contour_concept_captioning,scholar_multagent_empathetic_response,scholar_creatiparser,scholar_face_net,scholar_av_token_pruning,scholar_hierarchical_stance_distillation,scholar_emoverse,scholar_street_satellite_drone,scholar_bridging_subjectivity_captioning,scholar_histllm,scholar_emostyle,scholar_cervical_cytology}.
\section{Method}

\subsection{Overview}

\begin{figure*}[!t]
    \centering
    \includegraphics[
        width=\textwidth,
        trim=0.3in 3.3in 1.7in 1.2in,
        clip
    ]{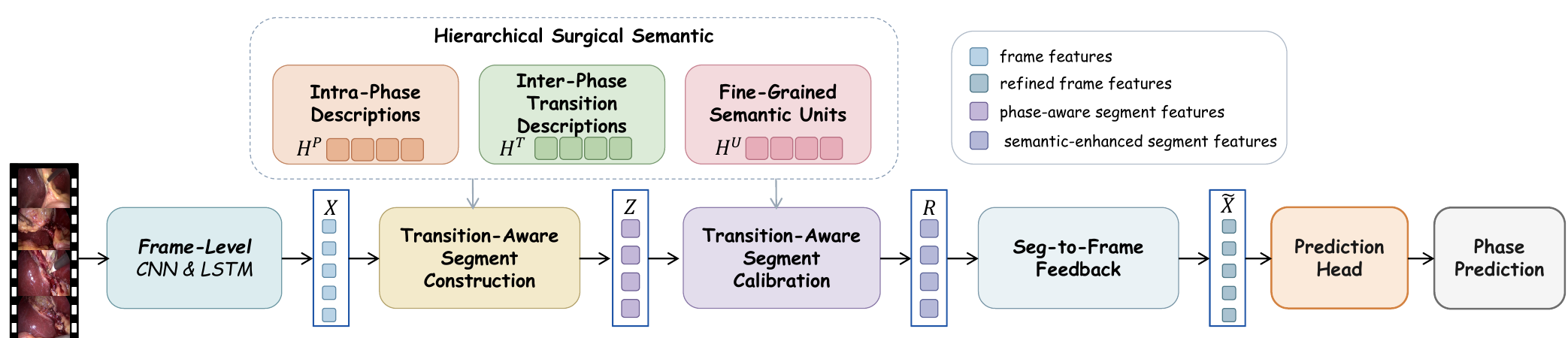}
    \caption{Overview of the proposed HTT-Net. The framework first extracts frame-level visual features and preliminary phase evidence from the input surgical video. A hierarchical surgical semantic memory is constructed from intra-phase descriptions, inter-phase transition descriptions, and fine-grained semantic units. TAS-Con constructs coherent phase-aware segment representations, which are further refined by TAS-Calib. Finally, the calibrated segment context is redistributed to frames for dense phase prediction.}
    \label{fig:framework}
\end{figure*}

Given a surgical video clip $\mathcal{V}=\{I_t\}_{t=1}^{T}$ with $T$ sampled frames, the goal of surgical video phase recognition is to predict a dense phase label sequence $\hat{\mathcal{Y}}=\{\hat{y}_t\}_{t=1}^{T}$. As illustrated in Fig.~\ref{fig:framework}, HTT-Net first extracts frame-level visual features $\mathbf{X}\in\mathbb{R}^{T\times d}$ and preliminary phase evidence $\mathbf{P}^{v}\in\mathbb{R}^{T\times C}$. A hierarchical surgical semantic memory is then constructed from textual descriptions at different granularities. Based on visual evidence and semantic memory, TAS-Con organizes frames into coherent phase-aware segments, and TAS-Calib further calibrates segment representations with hierarchical surgical semantics. The calibrated segment context is finally mapped back to frames for dense phase prediction.

\subsection{Hierarchical Surgical Textual Semantics}

We construct a hierarchical surgical semantic memory to provide structured procedural knowledge. Different from simple class-name prompts, the memory describes surgical procedures from three complementary levels: phase identity, transition plausibility, and fine-grained local discrimination.

Specifically, the memory contains intra-phase descriptions $\mathcal{D}^{P}$, inter-phase transition descriptions $\mathcal{D}^{T}$, and fine-grained semantic units $\mathcal{D}^{U}$. Intra-phase descriptions summarize the surgical objective, anatomical context, typical instruments, dominant actions, and visual cues within each phase. Inter-phase transition descriptions describe procedural changes between adjacent phases, such as changes in operative goals, tool usage, anatomical exposure, and local actions. Fine-grained semantic units describe reusable local cues, including instruments, anatomical structures, actions, and scene-level concepts.

A frozen text encoder $\phi_{\text{text}}(\cdot)$ is used to encode these descriptions into semantic embeddings:
\begin{equation}
\begin{aligned}
    \mathbf{H}^{P} &= \phi_{\text{text}}(\mathcal{D}^{P}), \\
    \mathbf{H}^{T} &= \phi_{\text{text}}(\mathcal{D}^{T}), \\
    \mathbf{H}^{U} &= \phi_{\text{text}}(\mathcal{D}^{U}).
\end{aligned}
\end{equation}
where $\mathbf{H}^{P}\in\mathbb{R}^{C\times d}$, $\mathbf{H}^{T}\in\mathbb{R}^{N_T\times d}$, and $\mathbf{H}^{U}\in\mathbb{R}^{N_U\times d}$ denote intra-phase, inter-phase transition, and fine-grained semantic-unit embeddings, respectively. The text encoder is kept frozen during training, and the encoded semantic memory is used as procedural reference knowledge for both TAS-Con and TAS-Calib.

\subsection{Transition-Aware Segment Construction}

\begin{figure*}[!t]
    \centering
    \includegraphics[
        width=0.78\textwidth,
        trim=1.55in 1.8in 0.95in 0.55in,
        clip
    ]{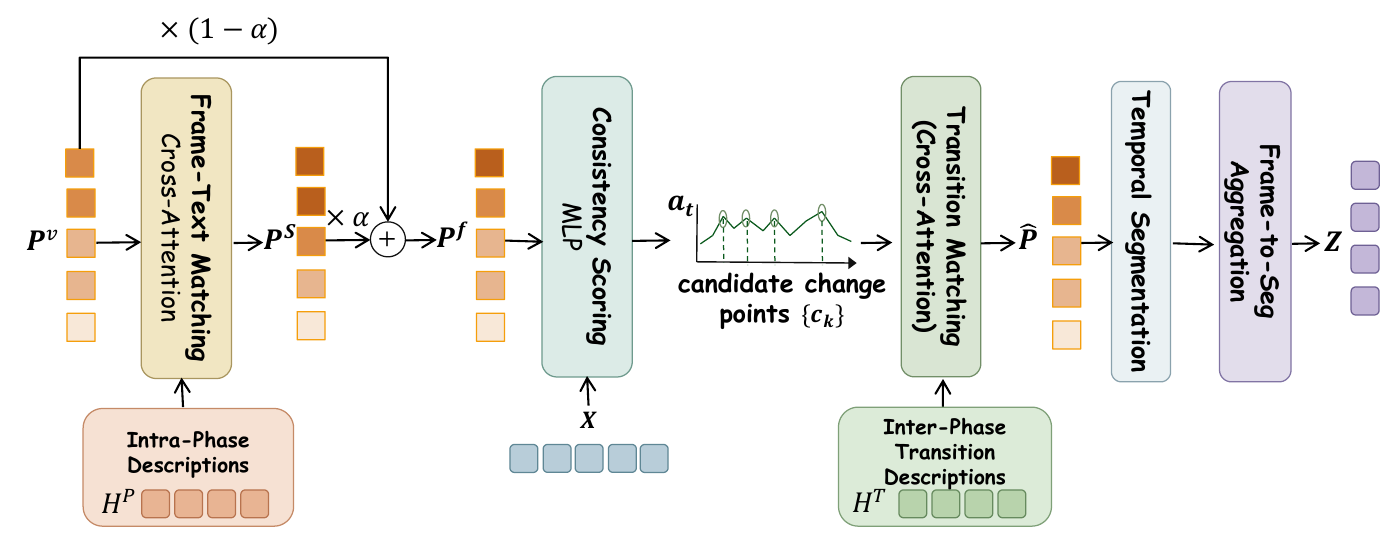}
    \vspace{-0.4em}
    {\footnotesize
    \caption{Illustration of Transition-Aware Segment Construction (TAS-Con), which constructs phase-aware segments using intra-phase semantics and inter-phase transition descriptions.}
    \label{fig:TAS-Con}
    }
    \vspace{-1.0em}
\end{figure*}

Frame-level predictions in surgical videos can be locally unstable, especially around phase boundaries. As shown in Fig.~\ref{fig:TAS-Con}, TAS-Con constructs phase-aware segments by combining preliminary phase evidence, visual context contrast, and transition semantics. Most sampled clips are dominated by a single phase and should remain phase-consistent, while boundary clips may contain frames from two adjacent phases. Directly relying on preliminary frame-wise predictions may therefore lead to over-segmentation, premature switching, delayed transitions, or short noisy segments. To address this issue, TAS-Con preserves phase-homogeneous clips as stable segments while allowing boundary clips to be divided into coherent adjacent phase segments when supported by both visual evidence and transition semantics.

The design of TAS-Con follows three steps. First, intra-phase textual semantics are used to stabilize preliminary frame-wise phase evidence, so that frames belonging to the same surgical phase are encouraged to maintain consistent phase responses. Second, visual context contrast and phase consistency cues are jointly used to locate candidate change points, which reduces the influence of isolated prediction noise. Third, inter-phase transition descriptions are used to verify whether a candidate change is procedurally plausible. In this way, TAS-Con does not simply split a clip whenever local visual evidence changes; instead, it checks whether the change is consistent with the expected evolution of surgical workflow.

Given frame-level features $\mathbf{X}=[\mathbf{x}_{1},\ldots,\mathbf{x}_{T}]$ and preliminary phase evidence $\mathbf{P}^{v}$, TAS-Con first introduces intra-phase semantics to obtain text-aware phase evidence.
The visual evidence is matched with intra-phase embeddings $\mathbf{H}^{P}$ to estimate a semantic phase prior $\mathbf{P}^{s}$, which is fused with the preliminary prediction:
\begin{equation}
\mathbf{P}^{f}=(1-\alpha)\mathbf{P}^{v}+\alpha\mathbf{P}^{s},
\end{equation}
where $\alpha$ controls the contribution of intra-phase semantics. The fused evidence provides a more stable phase estimate for subsequent segment construction.

Based on the fused phase evidence $\mathbf{P}^{f}$, TAS-Con then estimates candidate change points using both phase consistency and visual context contrast.
For each temporal position, local contexts before and after the position are compared, and a change score is computed by combining learned phase consistency cues with visual contrast cues.
High-response positions are selected as candidate change points.
For each candidate, TAS-Con further measures whether the local change is consistent with the inter-phase transition embeddings $\mathbf{H}^{T}$.
Specifically, the local transition query summarized from the left and right visual-phase contexts is compared with all inter-phase transition embeddings, and the maximum normalized similarity is used as the transition plausibility score.
Candidate points with high transition plausibility are retained to support phase boundary construction, while candidates with low plausibility are suppressed as transient visual variations.

This verification is important because many local visual variations in surgical videos are not true phase transitions. For example, tool motion, temporary occlusion, bleeding, or camera adjustment may cause abrupt appearance changes, but the underlying surgical objective may remain unchanged. By comparing local left-right contexts with inter-phase transition descriptions, TAS-Con can suppress such spurious changes while retaining transitions that correspond to meaningful procedural progress. Therefore, the generated segments are expected to be both visually coherent and semantically consistent.

The transition plausibility scores are used to refine local phase evidence around candidate positions. A locally smooth phase path is then decoded, and short noisy segments are merged. The resulting path divides the clip into $M$ temporal segments $\{S_m\}_{m=1}^{M}$. For each segment, frame-level features are aggregated into a segment representation:
\begin{equation}
    \mathbf{z}_{m}=\operatorname{Aggregate}
    \left(\{\mathbf{x}_{t}\mid t\in S_m\}\right), \quad
    \mathbf{Z}=[\mathbf{z}_{1},\ldots,\mathbf{z}_{M}].
\end{equation}
    
\subsection{Transition-Aware Segment Calibration}

\begin{figure*}[!t]
    \centering
    \includegraphics[
        width=0.8\textwidth,
        trim=1.3in 1.15in 0.2in 0.80in,
        clip
    ]{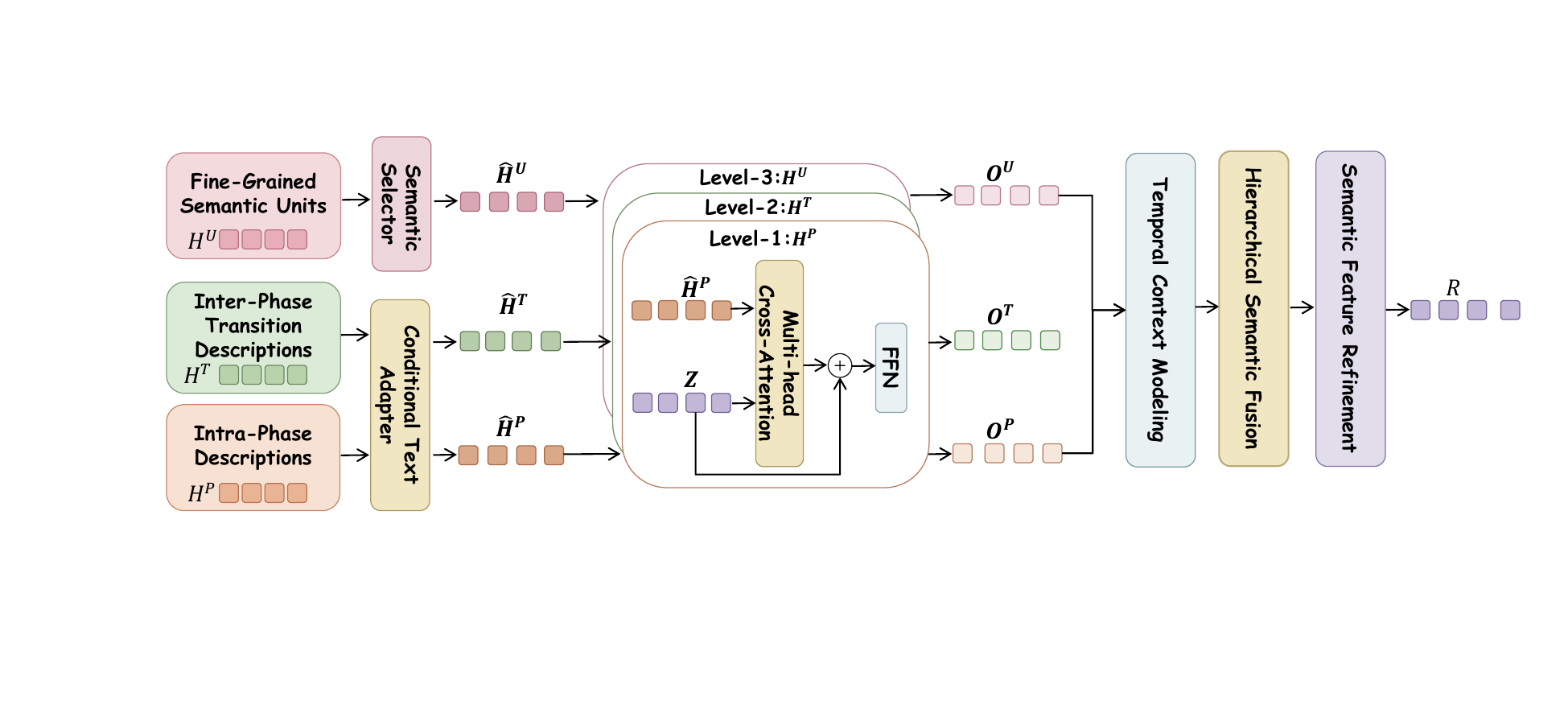}
    \vspace{-20pt}
    \caption{Illustration of Transition-Aware Segment Calibration (TAS-Calib), which refines phase-aware segment representations with hierarchical surgical semantic memory.}
    \label{fig:TAS-Calib}
    \vspace{-0.8em}
\end{figure*}

Although TAS-Con improves temporal coherence by constructing phase-aware segment representations, the resulting representations are still primarily derived from visual evidence.
When different phases share similar anatomical views, instruments, or actions, visual segment representations may remain ambiguous.
As illustrated in Fig.~\ref{fig:TAS-Calib}, TAS-Calib further enhances segment-level discrimination by calibrating segment representations through interaction with hierarchical surgical semantics.

Different semantic levels provide complementary guidance. Intra-phase semantics help identify the dominant procedural state of a segment, inter-phase transition semantics provide neighboring-phase context for ambiguous boundary segments, and fine-grained semantic units highlight local cues such as instruments, anatomical structures, actions, and scene concepts. Compared with dense frame-level vision-language fusion, TAS-Calib performs semantic interaction at the segment level, which substantially reduces computational cost while preserving the semantic information needed for phase discrimination.

Before cross-modal interaction, the semantic memory is adapted to the current clip. Intra-phase and inter-phase transition semantics are updated according to clip-level visual context, while fine-grained semantic units are selected according to their relevance to the current segment representations. The adapted semantic embeddings are computed as:
\begin{equation}
    \left(
    \widehat{\mathbf{H}}^{P},
    \widehat{\mathbf{H}}^{T},
    \widehat{\mathbf{H}}^{U}
    \right)
    =
    \operatorname{Adapt}
    \left(
    \mathbf{H}^{P},
    \mathbf{H}^{T},
    \mathbf{H}^{U},
    \mathbf{Z}
    \right).
\end{equation}
The adapted semantic embeddings provide clip-specific semantic references.

For each semantic level $k\in\{P,T,U\}$, TAS-Calib performs segment-to-text interaction through multi-head cross-attention:
\begin{equation}
    \mathbf{O}^{k}=\operatorname{MHCA}
    \left(\mathbf{Z},\widehat{\mathbf{H}}^{k},\widehat{\mathbf{H}}^{k}\right),
    \quad k\in\{P,T,U\}.
\end{equation}
Here, $\mathbf{O}^{P}$ emphasizes phase identity, $\mathbf{O}^{T}$ provides neighboring-phase transition context, and $\mathbf{O}^{U}$ supplies fine-grained local discrimination cues.

The multi-level semantic responses are then integrated with neighboring segment context by a hierarchical semantic fusion module:
\begin{equation}
    \mathbf{R}=\operatorname{Fuse}_{sem}
    \left(\mathbf{Z},\mathbf{O}^{P},\mathbf{O}^{T},\mathbf{O}^{U}\right).
\end{equation}
where $\mathbf{R}=[\mathbf{r}_{1},\ldots,\mathbf{r}_{M}]$ denotes the semantic-enhanced segment representations.

The fusion process adaptively balances visual segment evidence and semantic responses from different levels. For phase-homogeneous segments, intra-phase semantics can reinforce stable phase identity. For boundary segments, inter-phase transition semantics provide procedural context to reduce premature or delayed switching. Fine-grained semantic units further help distinguish visually similar phases by emphasizing local discriminative cues. The resulting semantic-enhanced segment representations are therefore more suitable for dense frame-level prediction after redistribution.

\subsection{Segment-to-Frame Prediction}

The calibrated segment representations are redistributed to frames according to the segment-to-frame correspondence obtained in TAS-Con. For each frame $t\in S_m$, the corresponding segment context $\mathbf{r}_{m}$ is assigned back to the frame and fused with the original frame-level feature. The final frame-level prediction is obtained by:
\begin{equation}
\begin{aligned}
    \widetilde{\mathbf{x}}_{t}
    &= \operatorname{Fuse}_{frm}(\mathbf{x}_{t},\mathbf{r}_{m}), \quad t\in S_m, \\
    \hat{\mathbf{p}}_{t}
    &= \operatorname{Softmax}
    \left(\operatorname{MLP}_{cls}(\widetilde{\mathbf{x}}_{t})\right).
\end{aligned}
\end{equation}
The final dense phase sequence is obtained by $\hat{y}_{t}=\arg\max_c \hat{p}_{t,c}$.

\subsection{Optimization}

The training objective follows a standard multi-level supervision strategy to support the proposed segment-based modeling framework. Specifically, frame-level phase supervision is used for dense prediction, segment-level supervision encourages discriminative segment representations, and boundary-related auxiliary supervision guides candidate change scoring in TAS-Con.

The frame-level classification loss provides the primary supervision for final phase prediction:
\begin{equation}
    \mathcal{L}_{cls}
    =
    -\frac{1}{T}\sum_{t=1}^{T}\log \hat{p}_{t,y_t}.
\end{equation}
where $y_t$ is the ground-truth phase label of frame $t$.

Since TAS-Con generates phase-aware segments without manual segment annotations, each segment is assigned a pseudo label according to the dominant frame-level label within the segment:
\begin{equation}
    y^{seg}_{m}
    =
    \operatorname{Majority}\left(\{y_t\mid t\in S_m\}\right).
\end{equation}
A segment classifier predicts the segment-level phase probability
$\hat{\mathbf{p}}^{seg}_{m}=\operatorname{Softmax}(\operatorname{MLP}_{seg}(\mathbf{r}_{m}))$
from the semantic-enhanced segment representation $\mathbf{r}_{m}$.
The segment-level loss is defined as:
\begin{equation}
    \mathcal{L}_{seg}
    =
    -\frac{1}{M}\sum_{m=1}^{M}
    \log \hat{p}^{seg}_{m,y^{seg}_{m}}.
\end{equation}
This loss encourages the calibrated segment representations to be discriminative at the procedural state level.

In addition, a boundary-related auxiliary loss is derived from frame-level phase changes to guide candidate change scoring:
\begin{equation}
    \mathcal{L}_{bnd}
    =
    -\frac{1}{T}\sum_{t=1}^{T}
    \left[
    g_t\log a_t+(1-g_t)\log(1-a_t)
    \right].
\end{equation}
where $g_t$ denotes the derived boundary label and $a_t$ denotes the predicted change response. This supervision encourages higher change responses near true phase transitions and lower responses inside phase-homogeneous intervals, helping TAS-Con avoid unnecessary segment splits while preserving meaningful transition structures.

The overall objective is:
\begin{equation}
    \mathcal{L}
    =
    \mathcal{L}_{cls}
    +
    \lambda_{seg}\mathcal{L}_{seg}
    +
    \lambda_{bnd}\mathcal{L}_{bnd}.
\end{equation}
where $\lambda_{seg}$ and $\lambda_{bnd}$ are weighting coefficients.

\section{Experiments}

\subsection{Experimental Setup}

\subsubsection{Datasets}

We evaluate the proposed HTT-Net on the Cholec80 and LCRS-100 datasets.

\textbf{Cholec80} is a widely used benchmark dataset for surgical video phase recognition~\cite{twinanda2016endonet,m2cai2016challenge}.
It contains 80 laparoscopic cholecystectomy videos annotated with 7 surgical phases.
Following the standard protocol used in previous studies~\cite{jin2017svrcnet,czempiel2020tecno,gao2021transsvnet,yang2024surgformer,zhang2026b2qnet}, the first 40 videos are used for training and the remaining 40 videos are used for testing.
All videos are sampled at 1 frames per second.

\textbf{LCRS-100} is an in-house laparoscopic colorectal resection surgery dataset containing 100 surgical videos with frame-level phase annotations and tool annotations.
It covers representative colorectal resection procedures, including right hemicolectomy, left hemicolectomy, sigmoid colectomy, and rectal resection.
The dataset is split into 45 training videos, 10 validation videos, and 45 testing videos.
Each frame is annotated with one of 23 surgical phases, and 10 tool categories are also provided.
Compared with Cholec80, LCRS-100 contains more phase categories, more diverse surgical workflows, and larger inter-case variations, making it more challenging for temporally coherent phase recognition.

\subsubsection{Evaluation Metrics}

Following previous surgical video phase recognition studies~\cite{twinanda2016endonet,jin2017svrcnet,czempiel2020tecno,gao2021transsvnet,yang2024surgformer,zhang2026b2qnet}, we report accuracy, precision, recall, and Jaccard index.
Accuracy is calculated by averaging per-video accuracy scores.
Since surgical phase distributions are usually imbalanced, precision, recall, and Jaccard index are computed for each phase and then averaged over all phases.
This protocol provides a more balanced evaluation for short or infrequent phases.

For Cholec80, we report both relaxed and unrelaxed results.
In the relaxed setting, predictions within a 10-second temporal window around phase boundaries are considered correct if they correspond to adjacent phases.
This setting reduces the penalty caused by annotation ambiguity near gradual phase transitions and follows the evaluation protocol adopted in prior work~\cite{twinanda2016endonet,jin2017svrcnet,zhang2026b2qnet}.
The unrelaxed setting directly evaluates frame-wise correctness without boundary tolerance.

\subsubsection{Implementation Details}

The proposed framework is implemented in PyTorch.
ConvNeXtV2-Tiny~\cite{woo2023convnextv2} is used as the visual backbone.
Input frames are resized to $384 \times 216$ and sampled at 1 fps.
Unless otherwise specified, the input sequence length is set to 256 frames.
The model is trained using AdamW with an initial learning rate of $1\times10^{-4}$ and a weight decay of $1\times10^{-2}$.

The hierarchical surgical semantic memory contains intra-phase descriptions, inter-phase transition descriptions, and fine-grained semantic units.
A frozen PubMedBERT encoder~\cite{gu2021pubmedbert} is leveraged to encode textual descriptions, and all text embeddings are pre-computed before training.
Therefore, no online text encoding is required during inference.
Additionally, our HTT-Net performs clip-based dense prediction by default.
We also report on the online frame-based variant HTT-Net-F, which only uses past and present information, to compare with online frame-based methods.

\subsection{Experimental Results}

\begin{table*}[!t]
\centering
\caption{Comparison results of HTT-Net with state-of-the-art methods on Cholec80. HTT-Net-F denotes the online frame-based variant that supports frame-wise inference using only past and present information. Standard deviations for some methods are unavailable due to missing reports. Our methods and the best results within each inference scheme are highlighted in bold.}
\label{tab:cholec80_results}
\setlength{\tabcolsep}{4.8pt}
\renewcommand{\arraystretch}{1.08}
\small
\begin{tabular}{c|c|lcccc}
\hline
Evaluation & Inference scheme & Methods & Accuracy & Precision & Recall & Jaccard \\
\hline
\multirow{16}{*}{Relaxed}
& \multirow{11}{*}{Frame-based}
& SV-RCNet~\cite{jin2017svrcnet} & $85.3\pm7.3$ & $80.7\pm7.0$ & $83.5\pm7.5$ & -- \\
& & TMRNet~\cite{jin2021tmrnet} & $90.1\pm7.6$ & $90.3\pm3.3$ & $89.5\pm5.0$ & $79.1\pm5.7$ \\
& & Trans-SVNet~\cite{gao2021transsvnet} & $90.3\pm7.1$ & $90.7\pm5.0$ & $88.8\pm7.4$ & $79.3\pm6.6$ \\
& & Not E2E~\cite{yi2022note2e} & $91.5\pm7.1$ & -- & $86.8\pm8.5$ & $77.2\pm11.2$ \\
& & UATD~\cite{ding2023less} & $91.9\pm5.6$ & $89.5\pm4.4$ & $90.5\pm5.9$ & $79.9\pm8.5$ \\
& & CMTNet~\cite{yue2023cmtnet} & $92.9\pm5.9$ & $90.1\pm7.1$ & $92.0\pm4.4$ & $81.5\pm10.4$ \\
& & LoViT~\cite{liu2025lovit} & $92.4\pm6.3$ & $89.9\pm6.1$ & $90.6\pm4.4$ & $81.2\pm9.1$ \\
& & SKiT~\cite{liu2023skit} & $93.4\pm5.2$ & $90.9$ & $91.8$ & $82.6$ \\
& & Surgformer~\cite{yang2024surgformer} & $93.4\pm6.4$ & $91.9\pm4.7$ & $92.1\pm5.8$ & $84.1\pm8.0$ \\
& & B2Q-Net-F~\cite{zhang2026b2qnet} & $95.3\pm6.5$ & $\mathbf{94.1\pm3.8}$ & $92.7\pm4.8$ & $86.5\pm8.3$ \\
& & \textbf{HTT-Net-F} & $\mathbf{95.9\pm4.6}$ & $94.0\pm3.9$ & $\mathbf{94.3\pm5.9}$ & $\mathbf{88.7\pm7.9}$ \\
\cline{2-7}
& \multirow{5}{*}{Clip-based}
& LTContext~\cite{bahrami2023longterm} & $92.7\pm7.9$ & $89.3\pm6.2$ & $90.1\pm7.9$ & $81.3\pm9.4$ \\
& & DiffAct~\cite{liu2023diffact} & $93.2\pm5.4$ & $90.0\pm5.9$ & $90.8\pm6.3$ & $82.5\pm8.9$ \\
& & FACT~\cite{lu2024fact} & $94.6\pm4.6$ & $92.8\pm4.1$ & $91.7\pm8.2$ & $85.8\pm9.5$ \\
& & B2Q-Net~\cite{zhang2026b2qnet} & $95.8\pm3.5$ & $\mathbf{93.9\pm4.0}$ & $92.0\pm7.0$ & $86.0\pm8.5$ \\
& & \textbf{HTT-Net} & $\mathbf{96.4\pm3.6}$ & $93.5\pm4.3$ & $\mathbf{93.7\pm4.4}$ & $\mathbf{88.3\pm7.5}$ \\
\cline{1-7}
\multirow{12}{*}{Unrelaxed}
& \multirow{7}{*}{Frame-based}
& Trans-SVNet~\cite{gao2021transsvnet} & $89.1\pm7.0$ & $84.7$ & $83.6$ & $72.5$ \\
& & LoViT~\cite{liu2025lovit} & $91.5\pm6.1$ & $83.1$ & $86.5$ & $74.2$ \\
& & SKiT~\cite{liu2023skit} & $92.5\pm5.1$ & $84.6$ & $88.5$ & $76.7$ \\
& & Surgformer~\cite{yang2024surgformer} & $92.4\pm6.4$ & $87.9\pm6.9$ & $89.3\pm7.8$ & $79.9\pm10.2$ \\
& & DACAT~\cite{yang2025dacat} & $93.5\pm5.3$ & $87.5\pm6.1$ & $88.5\pm8.3$ & $79.1\pm11.7$ \\
& & B2Q-Net-F~\cite{zhang2026b2qnet} & $\mathbf{94.8\pm4.6}$ & $\mathbf{90.3\pm7.1}$ & $\mathbf{90.8\pm8.2}$ & $\mathbf{82.8\pm9.5}$ \\
& & \textbf{HTT-Net-F} & $94.2\pm4.9$ & $88.4\pm8.1$ & $90.4\pm8.0$ & $81.4\pm11.7$ \\
\cline{2-7}
& \multirow{5}{*}{Clip-based}
& LTContext~\cite{bahrami2023longterm} & $91.6\pm7.4$ & $84.2\pm8.3$ & $86.3\pm9.1$ & $76.2\pm11.7$ \\
& & DiffAct~\cite{liu2023diffact} & $91.8\pm6.1$ & $87.6\pm7.7$ & $88.3\pm8.6$ & $79.3\pm11.5$ \\
& & FACT~\cite{lu2024fact} & $93.9\pm4.7$ & $89.4\pm7.1$ & $89.2\pm9.8$ & $80.9\pm11.2$ \\
& & B2Q-Net~\cite{zhang2026b2qnet} & $\mathbf{95.0\pm3.9}$ & $90.7\pm6.7$ & $90.9\pm7.8$ & $83.5\pm9.8$ \\
& & \textbf{HTT-Net} & $94.9\pm3.9$ & $\mathbf{90.8\pm5.8}$ & $\mathbf{91.3\pm5.6}$ & $\mathbf{84.0\pm9.4}$ \\
\hline
\end{tabular}
\end{table*}

\subsubsection{Comparison with State-of-the-Art Methods}

Tables~\ref{tab:cholec80_results} and~\ref{tab:lcrs_results} compare HTT-Net with existing methods.
On Cholec80, we report frame-based and clip-based methods under relaxed and unrelaxed evaluation, following recent surgical phase recognition studies.
On LCRS-100, representative baselines are reproduced under the same data split and training protocol.

\begin{figure*}[!t]
\centering
\includegraphics[
    width=\textwidth,
    trim=0in 1.4in 0.5in 0.05in,
    clip
]{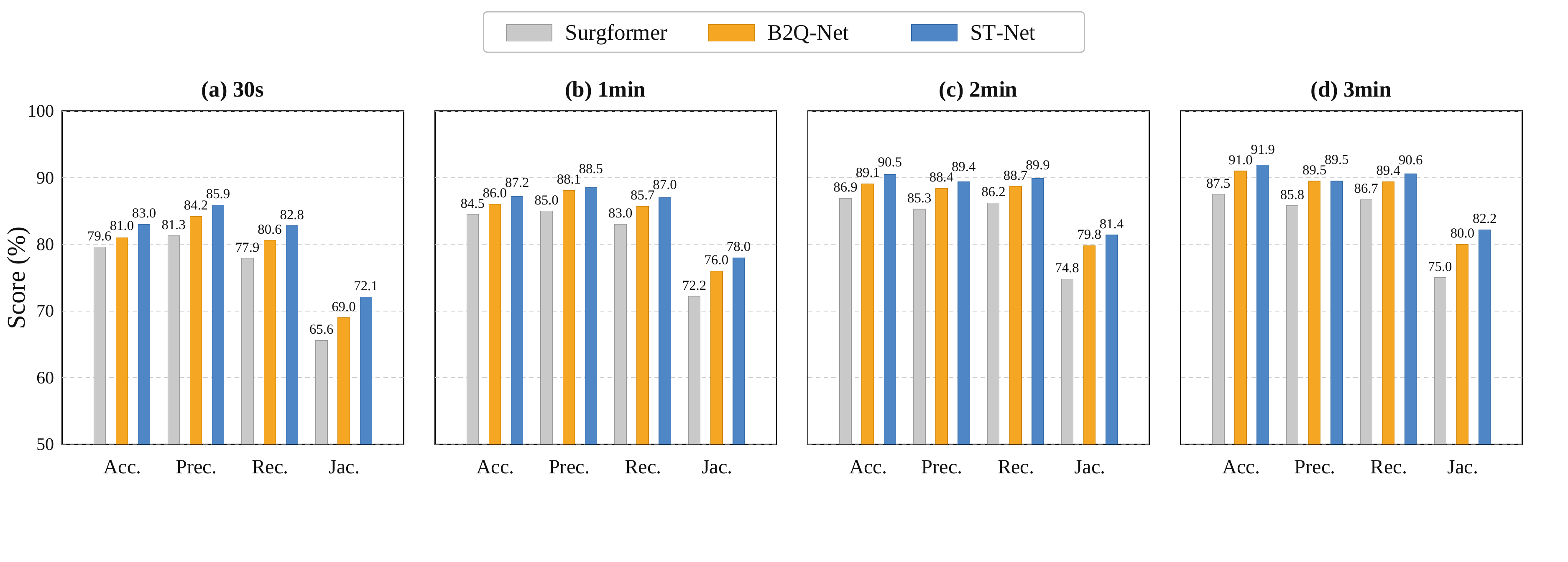}
\vspace{-0.6em}
\caption{Performance comparison within temporal windows around phase transitions on Cholec80. HTT-Net achieves consistently better performance than Surgformer and B2Q-Net across different window sizes, indicating stronger robustness around ambiguous phase boundaries.}
\label{fig:transition_window}
\end{figure*}

As shown in Table~\ref{tab:cholec80_results}, the proposed HTT-Net achieves strong performance on the Cholec80 dataset under both relaxed and unrelaxed evaluation settings.
Under relaxed evaluation, the clip-based HTT-Net obtains the best Accuracy, Recall, and Jaccard index among clip-based methods.
In particular, the Jaccard improvement over the compared B2Q-Net indicates that the proposed method not only improves frame-wise correctness but also produces more coherent phase regions.
Under unrelaxed evaluation, our HTT-Net achieves the best precision, recall, and Jaccard index among clip-based methods, while maintaining comparable accuracy to B2Q-Net.
These results suggest that semantic transition modeling helps reduce fragmented predictions and handle gradual phase changes.

HTT-Net-F also shows competitive frame-based performance.
In the relaxed setting, it achieves the best accuracy, recall, and Jaccard index within the frame-based group, demonstrating that the proposed modeling strategy can benefit online inference with only past and present information.
Since HTT-Net-F cannot access future frames, the clip-based HTT-Net is used as the main setting for evaluating the complete transition-aware segment modeling framework.

On LCRS-100, HTT-Net clearly outperforms the reproduced baselines.
The larger performance gap reflects the greater difficulty of LCRS-100, where more phases and diverse workflows introduce stronger visual ambiguity.
Many phases share similar tools, anatomical views, or local visual patterns, making purely visual temporal modeling insufficient.
By using hierarchical surgical semantic memory, HTT-Net calibrates ambiguous segment representations with intra-phase, inter-phase transition, and fine-grained semantic cues, leading to more reliable phase segments in complex colorectal procedures.

\begin{table}[!t]
\centering
\caption{Comparison results of HTT-Net with reproduced baselines on LCRS-100 under the unrelaxed setting. Accuracy denotes video-level accuracy. The best results within each inference scheme are highlighted in bold.}
\label{tab:lcrs_results}
\setlength{\tabcolsep}{2.8pt}
\renewcommand{\arraystretch}{1.05}
\small
\begin{tabular}{c|lcccc}
\hline
Inference scheme  & Method & Acc. & Prec. & Rec. & Jac. \\
\hline
\multirow{7}{*}{Frame-based}
& ResNet-50~\cite{he2016resnet} & $32.6$ & $18.3$ & $16.9$ & $9.2$ \\
& DenseNet-121~\cite{huang2017densely} & $36.2$ & $20.1$ & $16.0$ & $9.7$ \\
& TeCNO~\cite{czempiel2020tecno} & $24.7$ & $5.4$ & $8.3$ & $3.3$ \\
& Trans-SVNet~\cite{gao2021transsvnet} & $36.2$ & $13.6$ & $11.1$ & $6.6$ \\
& Surgformer~\cite{yang2024surgformer} & $37.0$ & $7.5$ & $6.9$ & $3.0$ \\
& Not E2E~\cite{yi2022note2e} & $40.0$ & $22.5$ & $\mathbf{22.6}$ & $13.8$ \\
& \textbf{HTT-Net-F} & $\mathbf{50.5}$ & $\mathbf{24.1}$ & $20.5$ & $\mathbf{14.0}$ \\
\hline
\multirow{5}{*}{Clip-based}
& LTContext~\cite{bahrami2023longterm} & $42.5$ & $17.7$ & $18.0$ & $9.6$ \\
& FACT~\cite{lu2024fact} & $43.6$ & $22.4$ & $24.6$ & $14.8$ \\
& DiffAct~\cite{liu2023diffact} & $45.9$ & $23.3$ & $24.0$ & $15.0$ \\
& B2Q-Net~\cite{zhang2026b2qnet} & $44.5$ & $19.6$ & $18.0$ & $12.0$ \\
& \textbf{HTT-Net} & $\mathbf{51.1}$ & $\mathbf{29.3}$ & $\mathbf{25.9}$ & $\mathbf{17.0}$ \\
\hline
\end{tabular}
\end{table}

\subsubsection{Analysis Around Phase Transitions}

Phase transitions are one of the most challenging parts of surgical video phase recognition.
Around boundaries, adjacent phases may share similar visual appearances, and the transition process is usually gradual rather than instantaneous.
To evaluate temporal robustness near phase boundaries, we compare different methods within temporal windows around annotated phase transitions on Cholec80.
Specifically, we report accuracy, precision, recall, and Jaccard index within 30-second, 1-minute, 2-minute, and 3-minute windows around phase transition points.

\begin{table}[!t]
\centering
\caption{Difficulty-aware phase analysis on Cholec80. Hard phases are defined as phases where B2Q-Net obtains phase-level Jaccard below 85\%.}
\label{tab:difficulty_phase_analysis}
\setlength{\tabcolsep}{4.5pt}
\renewcommand{\arraystretch}{1.12}
\small
\begin{tabular}{llcccc}
\hline
Subset & Method & Acc. & Prec. & Rec. & Jac. \\
\hline
\multirow{3}{*}{Easy phases} 
& Surgformer & $92.73$ & $89.66$ & $89.29$ & $81.47$ \\
& B2Q-Net    & $96.31$ & $\mathbf{97.88}$ & $94.94$ & $93.06$ \\
& HTT-Net     & $\mathbf{97.43}$ & $97.32$ & $\mathbf{96.86}$ & $\mathbf{94.37}$ \\
\hline
\multirow{3}{*}{Hard phases}
& Surgformer & $82.31$ & $80.01$ & $81.76$ & $67.85$ \\
& B2Q-Net    & $\mathbf{87.52}$ & $82.23$ & $87.09$ & $73.19$ \\
& HTT-Net     & $87.13$ & $\mathbf{85.86}$ & $\mathbf{87.14}$ & $\mathbf{76.17}$ \\
\hline
\end{tabular}
\end{table}

As shown in Fig.~\ref{fig:transition_window}, HTT-Net consistently outperforms the compared methods across different transition-window settings.
The improvement is particularly meaningful for Jaccard index, which is sensitive to fragmented predictions and boundary inconsistency.
This observation is consistent with the design motivation of TAS-Con.
Instead of directly trusting unstable frame-wise predictions near boundaries, TAS-Con constructs phase-aware segments by considering both visual context contrast and transition semantics.
This design helps suppress premature switching and short-term oscillations caused by transient visual changes.
Meanwhile, TAS-Calib further introduces hierarchical semantic references to distinguish adjacent phases with similar visual cues.
These two designs jointly improve recognition robustness around ambiguous phase transitions.

We further conduct a difficulty-aware phase analysis on Cholec80.
Hard phases are defined as phases where B2Q-Net obtains phase-level Jaccard below 85\%.
This analysis helps examine whether HTT-Net only improves easy and visually distinctive phases, or whether it also benefits ambiguous phase categories with stronger visual confusion.

As reported in Table~\ref{tab:difficulty_phase_analysis}, the proposed HTT-Net improves the Jaccard index on both easy and hard phase subsets.
For hard phases, our HTT-Net obtains the best Precision, Recall, and Jaccard index, although its Accuracy is slightly lower than the compared B2Q-Net.
This result suggests that our HTT-Net is more effective in reducing phase-level confusion and improving segment consistency for visually ambiguous phases.
The gain on hard phases is important because these phases usually contain shared instruments, similar anatomical backgrounds, or gradual procedural changes, where inter-phase transition descriptions and fine-grained semantic cues are expected to provide useful complementary guidance.

\begin{figure*}[!t]
\centering
\includegraphics[
    width=\textwidth,
    height=0.34\textheight,
    keepaspectratio
]{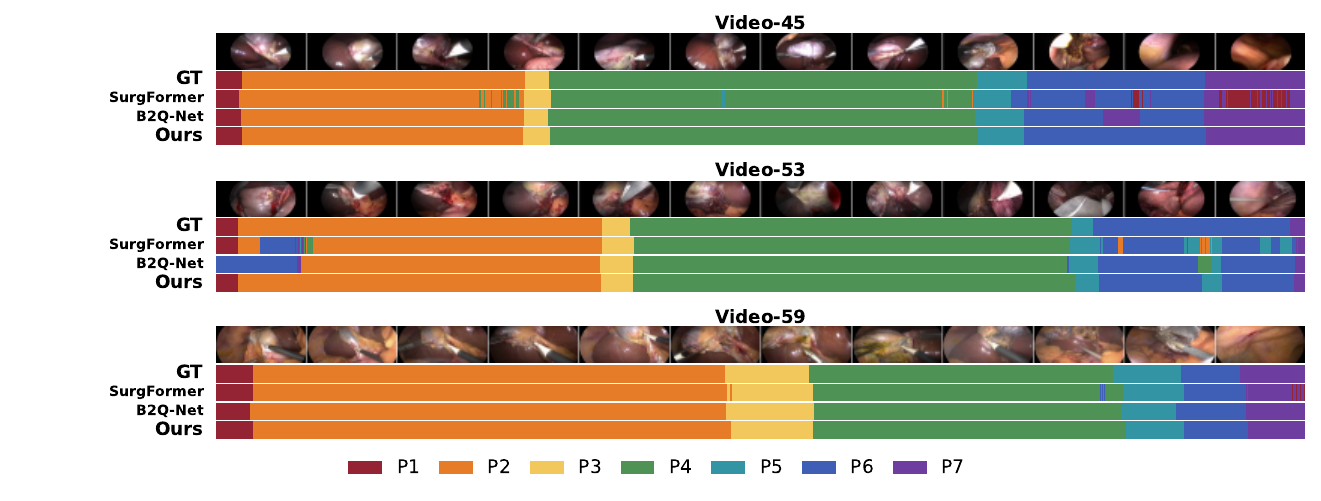}
\caption{Qualitative visualization of prediction timelines on Cholec80. The proposed method produces phase sequences that are highly consistent with ground truth and preserves clear phase boundaries.}
\label{fig:qualitative_timeline}
\end{figure*}

\subsubsection{Visualization Results}

We visualize representative prediction timelines on Cholec80 to examine the temporal behavior of HTT-Net.
Each timeline contains the ground-truth phase sequence and the predicted phase sequence.
Different colors denote different surgical phases. As illustrated in Fig.~\ref{fig:qualitative_timeline}, the predicted timelines are well aligned with the ground-truth phase sequences.
The proposed HTT-Net preserves the global procedural order and produces stable predictions within long phase intervals, which is important for reliable surgical workflow analysis.

\begin{figure}[!t]
    \centering
    \includegraphics[width=\linewidth]{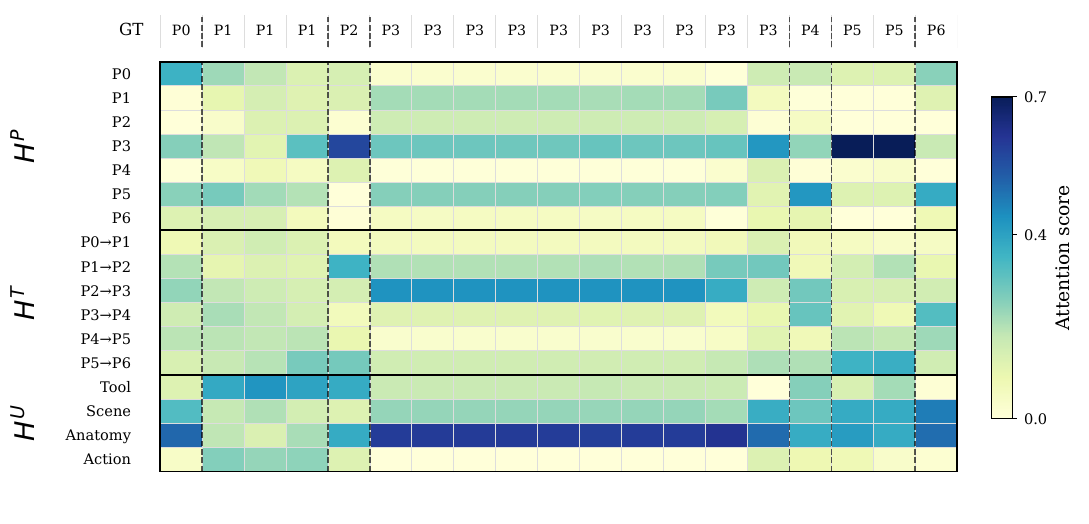}
    \caption{Semantic attention maps over hierarchical surgical semantic memory. Rows are grouped by intra-phase memory $H^{P}$, inter-phase transition memory $H^{T}$, and fine-grained memory $H^{U}$. The top row indicates the ground-truth phase of each segment, and dashed vertical lines denote phase changes.}
    \label{fig:semantic_attention}
\end{figure}

To further examine how hierarchical surgical textual semantics participate in segment-level calibration, we visualize the semantic attention responses between segment tokens and text embeddings.
As shown in Fig.~\ref{fig:semantic_attention}, the semantic responses are organized according to the three levels of surgical text descriptions.
The intra-phase text embeddings $H^{P}$ provide phase-related responses for stable procedural states.
The inter-phase transition text embeddings $H^{T}$ become informative around phase changes, where neighboring phase cues may coexist.
The fine-grained semantic-unit embeddings $H^{U}$ supply local concepts related to instruments, anatomy, and actions.
These complementary responses indicate that different text levels focus on different aspects of the surgical workflow, supporting segment-level feature calibration in TAS-Calib.

\subsection{Ablation Study}

\subsubsection{Effect of Model Components}

\begin{table}[!t]
\centering
\caption{Component-level ablation study of TAS-Con and TAS-Calib on LCRS-100.}
\label{tab:ablation_component}
\setlength{\tabcolsep}{5.5pt}
\renewcommand{\arraystretch}{1.08}
\begin{tabular}{cc|cccc}
\hline
TAS-Con & TAS-Calib & Accuracy & Precision & Recall & Jaccard \\
\hline
-- & -- & $47.3$ & $21.8$ & $18.6$ & $11.4$ \\
\checkmark & -- & $50.0$ & $28.0$ & $22.8$ & $14.7$ \\
-- & \checkmark & $50.7$ & $25.3$ & $24.4$ & $14.8$ \\
\checkmark & \checkmark & $\mathbf{51.1}$ & $\mathbf{29.3}$ & $\mathbf{25.9}$ & $\mathbf{17.0}$ \\
\hline
\end{tabular}
\end{table}

To verify the contribution of the proposed modules, we conduct component-level ablation experiments on LCRS-100.
The ablation focuses on TAS-Con and TAS-Calib.
TAS-Con evaluates whether Transition-Aware Segment Construction improves temporal organization, while TAS-Calib evaluates whether semantic calibration improves phase discrimination.

As shown in Table~\ref{tab:ablation_component}, both TAS-Con and TAS-Calib improve the baseline.
Adding TAS-Con improves accuracy from 47.3\% to 50.0\% and Jaccard index from 11.4\% to 14.7\%.
This indicates that TAS-Con helps suppress short-term noisy predictions and produces more coherent temporal segments.
Adding TAS-Calib also improves all metrics, especially recall and Jaccard index, showing that hierarchical surgical semantics provide useful discriminative cues for visually ambiguous phases.
When TAS-Con and TAS-Calib are combined, the model achieves the best overall performance.
This suggests that temporally coherent segment construction and semantic calibration are complementary: TAS-Con improves the organization of visual evidence, while TAS-Calib improves the semantic separability of segment representations.

\subsubsection{Effect of Hierarchical Semantic Memory}

We further evaluate the effect of semantic memory granularity by using one, two, and three levels of textual semantics.
The one-level setting uses intra-phase memory $H^{P}$ only.
The two-level setting adds transition memory $H^{T}$.
The full setting further incorporates fine-grained memory $H^{U}$.

\begin{table}[!t]
\centering
\caption{Ablation study of hierarchical surgical semantic memory on LCRS-100.}
\label{tab:ablation_text}
\setlength{\tabcolsep}{5.5pt}
\renewcommand{\arraystretch}{1.08}
\small
\begin{tabular}{ccc|cccc}
\hline
$H^{P}$ & $H^{T}$ & $H^{U}$ & Acc. & Prec. & Rec. & Jac. \\
\hline
\checkmark & -- & -- & $48.5$ & $25.7$ & $21.3$ & $14.4$ \\
\checkmark & \checkmark & -- & $49.5$ & $28.6$ & $25.4$ & $16.1$ \\
\checkmark & \checkmark & \checkmark & $\mathbf{51.1}$ & $\mathbf{29.3}$ & $\mathbf{25.9}$ & $\mathbf{17.0}$ \\
\hline
\end{tabular}
\end{table}

As shown in Table~\ref{tab:ablation_text}, using intra-phase memory alone already improves the baseline, indicating that textual descriptions of surgical objectives, anatomical context, instruments, and visual cues provide useful phase-level priors.
Adding transition memory further improves all metrics, especially recall and Jaccard index.
This demonstrates the importance of modeling procedural changes between adjacent phases rather than treating phases as isolated categories.
When fine-grained memory is added, the model achieves the best performance.
This result suggests that local semantic units related to instruments, anatomy, actions, and scene concepts provide complementary cues for distinguishing visually similar phases.

\subsection{Effect of Input Sequence Length}

We analyze the influence of input sequence length on LCRS-100.
Longer clips can provide richer temporal context for phase-aware segment construction, but they may also introduce redundant or weakly related frames.
\begin{table}[!t]
\centering
\caption{Effect of input sequence length on LCRS-100. Longer sequences provide richer temporal context but may introduce redundant frames and additional computation.}
\label{tab:seq_length}
\setlength{\tabcolsep}{8.8pt}
\renewcommand{\arraystretch}{1.0}
\small
\begin{tabular}{ccccc}
\hline
Seq. Len. & Acc. & Prec. & Rec. & Jac. \\
\hline
64  & $38.8$ & $19.4$ & $17.3$ & $11.2$ \\
128 & $46.1$ & $19.6$ & $20.6$ & $13.1$ \\
256 & $\mathbf{51.1}$ & $\mathbf{29.3}$ & $\mathbf{25.9}$ & $\mathbf{17.0}$ \\
512 & $49.4$ & $25.7$ & $24.7$ & $15.0$ \\
\hline
\end{tabular}
\end{table}
Table~\ref{tab:seq_length} reports the results with different input sequence lengths.
A short sequence of 64 frames provides limited temporal context, which is insufficient for recognizing slowly evolving phases or identifying transition patterns.
Increasing the sequence length to 128 improves performance by providing a broader temporal neighborhood.
The best overall performance is obtained with 256 frames, showing that sufficient temporal context is beneficial for constructing reliable phase-aware segments.
When the sequence length is further increased to 512, performance decreases.
This suggests that excessively long clips may introduce redundant frames and make segment construction more difficult.
Therefore, we use 256 frames as the default setting to balance recognition performance and computational cost.

\subsection{Efficiency Analysis}

Although HTT-Net introduces hierarchical surgical semantics, the text encoder is frozen and all text embeddings are pre-computed.
Therefore, the additional inference cost mainly comes from lightweight segment-level semantic calibration rather than repeated text encoding.
Since TAS-Calib operates on compact segment tokens instead of dense frame tokens, the computational overhead remains moderate.

\begin{table}[!t]
\centering
\caption{Comparison of inference efficiency and resource consumption. FPS is measured with batch size 1.}
\label{tab:efficiency}
\setlength{\tabcolsep}{5.5pt}
\renewcommand{\arraystretch}{1.15}
\begin{tabular}{lcccc}
\hline
Method & FPS & TFLOPs & Memory & Params \\
\hline
CNN+TCN & $340$ & $1.5$ & $9.5$ GB & $28$ M \\
LoViT~\cite{liu2025lovit} & $16$ & $22.5$ & $34.3$ GB & $263$ M \\
Surgformer~\cite{yang2024surgformer} & $33$ & $14.3$ & $26.7$ GB & $178$ M \\
B2Q-Net~\cite{zhang2026b2qnet} & $106$ & $3.7$ & $15.2$ GB & $205$ M \\
HTT-Net & $321$ & $3.7$ & $13.2$ GB & $351$ M \\
\hline
\end{tabular}
\end{table}

As shown in Table~\ref{tab:efficiency}, the proposed HTT-Net achieves 321 FPS, which is substantially faster than LoViT, Surgformer, and B2Q-Net under the same clip-based inference setting.
Compared with B2Q-Net, our HTT-Net has the same TFLOPs but achieves higher inference speed and lower GPU memory consumption.
Although HTT-Net contains more parameters due to the hierarchical surgical semantic memory and TAS-Calib, these parameters do not introduce heavy dense frame-level vision-language interaction.
Instead, semantic interaction is performed at the segment level, which keeps the computational cost efficient. Compared with the lightweight CNN+TCN, HTT-Net achieves substantially stronger recognition performance with acceptable additional cost.
Overall, the proposed HTT-Net provides a practical balance between recognition performance and inference efficiency.

\section{Conclusion}

In this paper, we proposed HTT-Net for hierarchical semantic transition modeling in surgical video phase recognition. By combining TAS-Con with TAS-Calib, the proposed framework introduces structured surgical semantic knowledge into temporally coherent phase modeling. The hierarchical surgical semantic memory provides intra-phase, inter-phase transition, and fine-grained semantic references, while boundary-related supervision guides candidate change scoring without manual segment annotations. Future work will explore more scalable semantic memory construction and validation on larger multi-center surgical datasets.
\bibliographystyle{IEEEtran}
\bibliography{refs-arXiv}

\end{document}